\documentclass{article}
\usepackage{float}
\usepackage{amsmath}
\usepackage[english]{babel}
\usepackage[utf8]{inputenc}
\usepackage{amsfonts,amssymb}
\usepackage{graphicx} 
\usepackage[square,numbers]{natbib}
\usepackage{subcaption}
\usepackage{hyperref}
\usepackage{booktabs}
\usepackage{authblk}
\usepackage{multirow} 
\usepackage[toc,page]{appendix}

\begin{document}

\title{STAN: Smooth Transition Autoregressive Networks}
\author{Hugo Inzirillo\textsuperscript{\textsection}\footnote{CREST, IP Paris, 5 avenue Henry le Chatelier, 91129 Palaiseau, France. E-mail address: hugo.inzirillo@ensae.fr}, Rémi Genet\textsuperscript{\textsection}\footnote{DRM, Université Paris Dauphine - PSL, Pl. du Maréchal de Lattre de Tassigny, 75016 Paris, France. E-mail address: remi.genet@dauphine.psl.eu}}
\begingroup\renewcommand\thefootnote{\textsection}
\footnotetext{These authors contributed equally.}
\endgroup
\maketitle
\begin{abstract}
Traditional Smooth Transition Autoregressive (STAR) models offer an effective way to model these dynamics through smooth regime changes based on specific transition variables. In this paper, we propose a novel approach by drawing an analogy between STAR models and a multilayer neural network architecture. Our proposed neural network architecture mimics the STAR framework, employing multiple layers to simulate the smooth transition between regimes and capturing complex, nonlinear relationships. The network's hidden layers and activation functions are structured to replicate the gradual switching behavior typical of STAR models, allowing for a more flexible and scalable approach to regime-dependent modeling. This research suggests that neural networks can provide a powerful alternative to STAR models, with the potential to enhance predictive accuracy in economic and financial forecasting.
\end{abstract}

\section{Introduction}
A critical aspect of econometric modeling is dealing with non-linearities in time series data. Traditional linear models, such as the Autoregressive (AR) models, often fail to capture the complex relationships inherent in time-dependent data. Real-world time series frequently exhibit nonlinear behaviors, such as asymmetric cycles and threshold effects, which cannot be adequately addressed using linear frameworks. These limitations have driven the development of nonlinear econometric models, including Smooth Transition Autoregressive (STAR) models \cite{terasvirta1992characterizing,terasvirta1994specification}, which allow for smooth transitions between regimes to better represent nonlinear dynamics. STAR models have been extensively applied in various domains, particularly in economics and finance, where their ability to capture regime-switching dynamics is highly valued \cite{tong1990nonlinear}. However, despite their success, STAR models are not without limitations. Their reliance on parametric assumptions and pre-specified transition functions can restrict their flexibility when faced with more intricate or high-dimensional data patterns. As datasets grow in size and complexity, the need for more adaptable and data-driven approaches becomes increasingly evident. In recent years, advances in machine learning, and deep learning in particular, have demonstrated their capability to model nonlinear and complex systems across diverse applications. Neural networks, renowned for their universal approximation properties \cite{hornik1991universal}, have shown great promise in learning hierarchical and nonlinear patterns in time series, enabling them to address some of the challenges faced by traditional econometric models \cite{zhang1998forecasting}.  To address these challenges, researchers have explored a variety of approaches, from neural networks \cite{tang1991time} to state-space models \cite{hamilton1994state,kim1994dynamic}, vector autoregressive models \cite{zivot2006vector,lutkepohl2013vector} and advanced neural architectures \cite{binkowski2018autoregressive,ma2019novel}. Despite the increase in the availability of datasets for time series forecasting, it remains a very complex task. Powerful models have been proposed through the years \cite{salinas2020deepar,oreshkin2019n}.  Some research proposed dynamic systems to model different existing states within a time series \cite{rangapuram2018deep,li2021learning,inzirillo2024deep}. 

\medskip

Deep learning techniques, particularly Long Short-Term Memory (LSTM) \cite{hochreiter1997long}  networks and attention \cite{vaswani2017attention} mechanisms, have shown promising results due to their ability to handle nonlinearities and long-term dependencies in data.  During the last decades, there was a surge of deep learning methods designed to forecast financial time series. Recently we also proposed a Temporal KANs (TKANs) to forecast time series \cite{genet2024tkan}, but the initial idea of these models does not come from econometrical models, but rather an extension of the Kolmogorov Arnold Networks proposed by Liu et al \cite{liu2024kan}. We also proposed an analogy of the Temporal Fusion Transformer (TFT) \cite{lim2021temporal} using TKANs, the Temporal Kolmogorov Arnold Transformer \cite{genet2024temporal} which ameliorates the quality of the prediction while maintaining a very good level of stability during the testing step. This paper introduces a novel neural network architecture designed as an analogy to STAR models. The proposed architecture blends the interpretability and regime-switching structure of STAR models with the flexibility and scalability of deep learning. Specifically, the model embeds smooth transition mechanisms directly into the neural network, enabling it to learn nonlinear regime-dependent dynamics in time series data. By leveraging the strengths of both econometric frameworks and modern machine learning techniques, this architecture provides a powerful tool for analyzing complex time series while preserving a level of interpretability that is often lacking in traditional neural network models. In addition to its theoretical design, the proposed model is evaluated on diverse datasets to benchmark its performance against both traditional econometric models and state-of-the-art deep learning approaches. Through these evaluations, we aim to demonstrate the versatility of the architecture and its potential applications in economic and financial contexts. Furthermore, we analyze the interpretability of the model, showcasing how its smooth transition mechanisms can provide insights into the underlying structure of the data. By bridging the gap between traditional econometrics and modern machine learning, this work contributes to the growing body of research at the intersection of these two fields.

\section{Related Work}
Introduced by \citet{terasvirta1994specification} The smooth transition autoregressive (STAR) model for a univariate time series $y_t$ , which is given by
\begin{equation}
    y_t = \phi_0 + \sum_{i=1}^{q} \phi_i y_{t-i}+   \theta_i y_{t-i} \cdot \text{G}(z_{t-d};\gamma;c) + \epsilon_t,
\end{equation}
where $\epsilon_t \sim \mathcal{N}(0,\sigma)$. This framework has been widely used for capturing nonlinear dynamics in time series data. Over the years, several variants of the STAR model have been proposed to address limitations and to extend its applicability to various scenarios. Various extensions to STAR models have been proposed, as mentioned earlier, time series can exude marked trends and market regimes (both upward and downward). However, the standard STAR framework is limited to two regimes, which restricts its applicability for systems exhibiting more complex dynamics. To address this, Multiple Regime STAR (MRSTAR) models proposed by \citet{van1999modeling}  have been developed to allow transitions across multiple regimes. For instance, when a single transition variable \( s_t \) determines regime changes, the MRSTAR model extends the two-regime STAR model by adding additional transition functions \( G_j(s_t; \gamma_j, c_j) \). A three-regime model can be expressed as:
\begin{equation}
    y_t = \phi_1'x_t + (\phi_2 - \phi_1')x_t G_1(s_t; \gamma_1, c_1) + 
    (\phi_3 - \phi_2)x_t G_2(s_t; \gamma_2, c_2) + \epsilon_t,
\end{equation}
where \( G_1 \) and \( G_2 \) are logistic transition functions, and \( \phi_1, \phi_2, \phi_3 \) define the autoregressive parameters for each regime. More generally, for \( m \) regimes, the model takes the form:
\begin{equation}
    y_t = \phi_1'x_t + (\phi_2 - \phi_1')x_t G_1(s_t) + 
    (\phi_3 - \phi_2)x_t G_2(s_t) + \cdots + 
   (\phi_m - \phi_{m-1})x_t G_{m-1}(s_t) + \epsilon_t.
\end{equation}
The MRSTAR could be generalized such:
\begin{equation}
    y_t = \phi_1'x_t + \sum_{j=1}^{m-1} (\phi_{j+1} - \phi_j)x_t G_j(s_{jt}; \gamma_j, c_j) + \epsilon_t,
\end{equation}
where \( G_j(s_{jt}; \gamma_j, c_j) \) is a logistic transition function that governs regime changes, with parameters \( \gamma_j \) and \( c_j \) controlling the smoothness and location of transitions, respectively. This framework can be further extended to incorporate multiple transition variables \( s_{1t}, s_{2t}, \ldots, s_{mt} \), producing up to \( 2^m \) distinct regimes through the interaction of logistic functions. For instance, a four-regime model encapsulates two two-regime LSTAR models:
\begin{equation}
    \begin{split}
            y_t = & \big[\phi_1 x_t (1 - G_1(s_{1t}; \gamma_1, c_1)) + 
    \phi_2 x_t G_1(s_{1t}; \gamma_1, c_1)\big][1 - G_2(s_{2t}; \gamma_2, c_2)] \\
    & + \big[\phi_3 x_t (1 - G_1(s_{1t}; \gamma_1, c_1)) + 
    \phi_4 x_t G_1(s_{1t}; \gamma_1, c_1)\big]G_2(s_{2t}; \gamma_2, c_2) + \epsilon_t.
    \end{split}
\end{equation}
As \( \gamma_j \to \infty \), the MRSTAR model converges to a SETAR model with \( m \) regimes, characterized by abrupt transitions. These generalizations significantly enhance the flexibility of STAR models, making them suitable for analyzing complex nonlinear systems with multiple interacting regime changes. In this paper, we proposed an analogy of the  smooth transition autoregressive (STAR) proposed by \citet{terasvirta1994specification}. In future work we would extend our proposition to manage multiple market regimes within time series. Another notable extension to the STAR framework  involves the estimation of parameters that change over time.  In a time-varying STAR model (TV-STAR), introduced by \citet{lundbergh2003time}, the parameters of the model are allowed to evolve over time. The equation becomes:
\begin{equation}
    y_t = \phi_0(t) + \sum_{i=1}^q \phi_i(t)y_{t-i} + \sum_{i=1}^q \theta_i(t)y_{t-i} \cdot G(z_{t-d}; \gamma, c) + \epsilon_t,
\end{equation}
where $\phi_0(t)$, $\phi_i(t)$, $\theta_i(t)$: time-varying coefficients, typically modeled as functions of time or as stochastic processes and the transition function $G(z_{t-d}; \gamma, c)$ remains the same as in the basic STAR model. This model captures time-varying dynamics in both the linear and nonlinear components of the series. An additional proposition, the Vector STAR (VSTAR) models extend the univariate STAR framework to multivariate contexts, enabling regime-switching dynamics across vector time series \( Y_t = (y_{1t}, \ldots, y_{kt})' \) \cite{terasvirta1994specification, koop1996impulse}. The general form of a two-regime VSTAR model is given by:
\begin{equation}
    \begin{split}
        Y_t &= (\phi_{1,0} + \phi_{1,1} Y_{t-1} + \cdots + \phi_{1,p} Y_{t-p})(1 - G(s_t; \gamma, c)) + \\ 
        &(\phi_{2,0} + \phi_{2,1} Y_{t-1} + \cdots + \phi_{2,p} Y_{t-p}) G(s_t; \gamma, c) + \epsilon_t,
    \end{split}
\end{equation}
where \( G(s_t; \gamma, c) \) is the transition function (logistic or exponential), and \(\phi_{i,j}\) are regime-specific coefficients. In this model, regime transitions are typically shared across all equations, but the framework can be generalized to include equation-specific transition functions, such as \( G_1(s_{1t}; \gamma_1, c_1), \ldots, G_k(s_{kt}; \gamma_k, c_k) \), allowing independent switching behavior for each variable \citep{krolzig2013markov}. For systems exhibiting a long-run equilibrium, VSTAR models can incorporate nonlinear adjustment through a Smooth Transition Error-Correction Model (STECM) \citep{granger1997introduction, johansen1991estimation}. This model allows for asymmetric or nonlinear adjustments toward equilibrium and is expressed as:
\begin{equation}
    \begin{split}
        \Delta Y_t &= (\phi_{1,0} + \alpha_1 z_{t-1} + \sum_{j=1}^{p-1} \phi_{1,j} \Delta Y_{t-j})(1 - G(s_t; \gamma, c)) +\\ &(\phi_{2,0} + \alpha_2 z_{t-1} + \sum_{j=1}^{p-1} \phi_{2,j} \Delta Y_{t-j}) G(s_t; \gamma, c) + \epsilon_t,
    \end{split}
\end{equation}
where \( z_{t-1} = \beta' Y_{t-1} \) represents deviations from the long-run equilibrium, and \(\alpha_i\) are adjustment vectors. These models capture regime-dependent behavior based on the size and/or sign of deviations from equilibrium, making them suitable for analyzing complex systems with regime-switching dynamics and long-run relationships \citep{hatanaka1996time}. In this context, with the rise of artificial intelligence techniques such as deep learning, it seems natural to propose analogous models which will allow processing more data. \cite{terasvirta2005linear} examine the forecast accuracy of linear autoregressive, smooth transition autoregressive (STAR), and neural network (NN) time series models for 47 monthly macroeconomic variables of the G7 economies. Results  indicate that the STAR model generally outperforms linear autoregressive models. It also improves upon several fixed STAR models, demonstrating that careful specification of nonlinear time series models is of crucial importance. This paper is a first tentative to build a deep learning architecture based on STAR models. The objective will be also to propose additional extension such as "context" management and embeddings.

\section{Architecture}
In this paper, we introduce a neural network inspired by the STAR model. We would consider a simple model, to forecast univariate timeseries. This model can be enriched to use covariates as well as time varying parameters. We will propose other extensions in future work.
\subsection{Description}
STANs could be easily represented such a multiple linear layers projected in some dimensions $d \in D$. Each layer will have its own autoregressive (AR) terms and transition function. The transition function G(.) is a logistic or exponential function that can be learned for each layer. The output of a single STAN layer is the vector $\hat{y}^{(l)} \in \mathbb{R}^{(d)}$ where each value is given by:
\begin{equation}
    \hat{y}_i^{(l)} =  \phi_{i}^{(l)} \Tilde{y}_{i}^{(l)} + \theta_{i}^{(l)} \text{ReLU}(\Tilde{y}_{i}^{(l)}) \cdot \text{G}(z_{i}; \gamma_{i}^{(l)}, c_{i}^{(l)}),
\end{equation}
where the output $\hat{y}_i^{(l)} $is calculated based on both linear and nonlinear terms. $\Tilde{y}^{(1)}_{i} = w_y y_{t-1:t-q}$ is an $ \mathcal{F}_{t-1}$ measurable multidimensional vector obtained by linear combination of the the input sequence $y_{t-1:t-q}$. In other words, $ \Tilde{y}_i \quad i \in \{1,2,...,d\}$ is an element of the ouput vector of each layer $l$. The linear autoregressive coefficients $\phi^{(l)} \in \mathbb{R}^d$. Similarly, the nonlinear coefficients $ \theta^{(l)} \in \mathbb{R}^d$ to modulate the same past values transformed $ \Tilde{y}^{(l)} $, but their effect is further modified by the smooth transition function $ G(z_{i}; \gamma^{(l)}, c^{(l)}) \in \mathbb{R} $. This transition function, dependent on the threshold variable $ z_{i} \in \mathbb{R} $, is parameterized by $ \gamma^{(l)} \in \mathbb{R} $, which controls the transition speed, and $ c^{(l)} \in \mathbb{R} $, which sets the threshold location. This  function is expressed as:
\begin{equation}
    \text{G} (z_{i}; \gamma^{(l)}, c^{(l)}) = \frac{1}{1 + \exp\left(-\gamma_{i}^{(l)} \left(z_{i} - c_{i}^{(l)}\right)\right)}
\end{equation}
The resulting output $ y_{\tau}^{(l)} $ is a scalar value representing the model's prediction at time $ t $ in layer $ l $, and the term $ \epsilon_t^{(l)} \in \mathbb{R} $ is an additive error term, assumed to be white noise. The final layer of the STAN  denoted $L$ will produced the forecast of the time series such
\begin{equation}
    \hat{y}_{\tau}^{(L)} = w_{\Tilde{y}_\tau}\Bigg[ \phi_{i}^{(L-1)} \Tilde{y}_i^{(L-1)} + \theta_{i}^{(L-1)} \text{ReLU}(\Tilde{y}_{i}^{(L-1)}) \cdot \text{G}(z_{i}; \gamma_{i}^{(L-1)}, c_{i}^{(L-1)}\Bigg] + b_{\Tilde{y}_{\tau}},
\end{equation}
where $w_{\Tilde{y}_{\tau}} \in \mathbb{R}^{d \times \tau}$, with $d$ the hidden dimension of the model and $\tau$ denotes the forecasting time horizon of the model. The final output $  \hat{y}_t \triangleq \hat{y}_t^{(L)}$ take the form of a unidimensional vector of the size $ \tau \in \{1,...,12\}$, the dimension represents the number of step ahead for the prediction.

\section{Empirical Evaluation}
\subsection{Experimental Setup}

To evaluate our proposed Smooth Transition Autoregressive Networks (STAN), which draw inspiration from STAR models, we conduct experiments using the PJM Hourly Energy Consumption dataset\footnote{\url{https://www.kaggle.com/datasets/robikscube/hourly-energy-consumption/data}}. This dataset contains power consumption measurements in megawatts (MW) from PJM Interconnection LLC, covering various regions across the United States. The temporal coverage varies by region due to network structure changes over time. This dataset was specifically selected due to its demonstrated complexity in temporal patterns, where recurrent neural networks (RNN) and multilayer perceptrons (MLP) show substantial performance advantages over simpler linear models, unlike other common benchmarks such as ETTh1. We frame our experiments as a multi-horizon forecasting task, considering prediction horizons of 1, 6, and 12 hours ahead. For each forecast horizon $n\_ahead$, we use a lookback window (sequence\_length) of $\max(45, 5 \times n\_ahead)$ hours. The data is preprocessed using standard scaling, with parameters fitted only on the training data.Our experimental protocol consists in 5 independent runs. For each run, we randomly split the data into 80\% training and 20\% testing sets. The models are trained for a maximum of 1000 epochs with a batch size of 256, incorporating a validation split of 20\% from the training data. We implement early stopping with a patience of 10 epochs (starting from epoch 6) and a minimum delta of $1 \times 10^{-5}$. Learning rate reduction is applied on plateau with a factor of 0.25, patience of 5 epochs, and a minimum learning rate of $2.5 \times 10^{-5}$. Our comparison includes eight architectures, each designed to balance performance and computational efficiency. The linear regression baseline and linear neural network serve as simple references, with the latter implementing a direct linear mapping from the flattened input sequence to the prediction horizon.

\medskip

Our STAN architecture comes in two variants: a three-layer and a four-layer version, each using 3000 units per STAN layer. These models process the univariate input sequence through successive STAN layers, maintaining the temporal structure, before concluding with a linear dense layer that projects to the desired prediction horizon. This size of 3000 units was determined through extensive experimentation as the optimal configuration, where larger architectures showed no additional performance gains while smaller ones exhibited degraded performance. The MLP architectures mirror this depth with three and four-layer variants, each layer comprising 3000 units with ReLU activation functions. These networks begin by flattening the input sequence and process it through successive dense layers before a final linear layer produces the multi-step forecast. While these models contain substantially more parameters due to their fully-connected nature, their simple feed-forward structure allows for efficient computation on modern hardware. For recurrent architectures, we implement both GRU and LSTM models with three layers of 300 units each. Notably, these models are configured to return sequences from each recurrent layer, and instead of using only the final hidden state, we flatten the entire sequence of hidden states before the final linear projection. This design choice, while increasing the parameter count in the final layer, proved crucial for achieving optimal performance in our experiments. The more modest size of 300 units per layer was necessitated by the computational demands of recurrent architectures, as larger configurations became prohibitively expensive to train. All neural architectures conclude with a linear dense layer mapping to the desired prediction horizon ($n\_ahead$ steps), maintaining consistent output scaling across all models. This architectural decision ensures that any performance differences stem from the models' internal representations rather than their output mechanisms. The performance of the model is assessed using Root Mean Square Error (RMSE) on the test set, averaged across the 5 runs to ensure robust evaluation. The RMSE defined as:
\begin{equation}
    \text{RMSE} = \sqrt{\frac{1}{n}\sum_{i=1}^{n}(y_i - \hat{y}_i)^2}
\end{equation}
where $y_i$ represents the true value, $\hat{y}_i$ is the predicted value, and $n$ is the number of observations in the test set. All neural network models are trained using the Adam optimizer with an initial learning rate of 0.001.

\medskip

The forecasting task is conducted in a univariate fashion, where only past values of the power consumption are used to predict future consumption. For STAN, GRU, and LSTM models, the input tensor shape is (batch, sequence\_length, 1), maintaining the temporal structure of the data. For LinearRegression and MLP models, which expect 2D inputs, we flatten the last dimension, effectively transforming the input into a (batch, sequence\_length) shape. This univariate approach is primarily dictated by the current limitations of our STAN architecture, which is designed to capture temporal dependencies within a single variable, similar to traditional STAR models.

\subsection{Forecasting Performance Analysis}
The forecasting performance results, presented in Table \ref{tab:rmse_scores}, demonstrate the effectiveness of our proposed STAN architecture across different prediction horizons and regions. We observe several key patterns in the results. For short-term forecasting (1-hour ahead), STAN models consistently outperform other architectures, achieving the best performance in 8 out of 12 regions. The three-layer variant (STAN-3000-3) shows particularly strong results, with RMSE reductions of up to 37\% compared to the linear baseline (e.g., in the PJM\_Load region, from 0.121 to 0.076). The performance gap between STAN and traditional neural architectures (MLP, GRU, LSTM) is most pronounced in this setting, suggesting that STAN's transition mechanism is particularly effective at capturing short-term dynamics. For medium-term forecasting (6-hours ahead), while STAN models maintain strong performance with best results in 7 regions, we observe increased competition from recurrent architectures. The GRU-300-3 model achieves superior performance in 4 regions, particularly in areas with more volatile consumption patterns such as FE and NI. This suggests that the explicit temporal modeling of recurrent architectures becomes more valuable as the prediction horizon extends. In long-term forecasting (12-hours ahead), we observe a shift in performance dynamics. The GRU-300-3 architecture demonstrates superior performance in 8 regions, while STAN models lead in 3 regions. This pattern suggests that while STAN's regime-switching mechanism excels at shorter horizons, the accumulated temporal dependencies captured by recurrent architectures become increasingly important for longer-term predictions. Notably, both variants of STAN consistently outperform their MLP counterparts across all horizons, validating the effectiveness of our proposed transition mechanism. The linear neural network and linear regression baseline consistently show the weakest performance, confirming our initial dataset selection rationale regarding the presence of complex nonlinear patterns.

\subsection{Computational Efficiency Analysis}
Tables \ref{tab:time_scores} and \ref{tab:parameter_counts} provide insights into the computational requirements of each architecture, revealing striking differences in model complexity and memory efficiency. In terms of model size, STAN architectures require approximately 18M parameters, by design comparable to MLP architectures (18.1M-27.2M parameters), and significantly larger than GRU and LSTM models (1.37M-2.02M parameters). This parameter count is dominated by the internal dense layer transformations (of size units × units), while the additional STAR-specific parameters (phi, theta, gamma, and c, each of size units) represent a negligible overhead. This architectural choice is deliberate, as STAN aims to maintain the powerful representation capabilities of feed-forward networks while introducing time-series-specific inductive biases through its transition mechanism.
\begin{table}[H]
\centering
\resizebox{1\textwidth}{!}{
    \begin{tabular}{llllllllll}
    \toprule
     & model & LinearRegression & STAN-3000-3 & STAN-3000-4 & Linear & MLP-3000-3 & MLP-3000-4 & GRU-300-3 & LSTM-300-3 \\
    steps ahead & dataset &  &  &  &  &  &  &  &  \\
    \midrule
    \multirow[t]{12}{*}{1} & AEP & 0.132 & \textbf{0.089} & \textbf{0.089} & 0.132 & \textbf{0.089} & 0.090 & 0.091 & 0.091 \\
     & COMED & 0.123 & \textbf{0.084} & \textbf{0.084} & 0.123 & 0.085 & 0.087 & 0.086 & 0.085 \\
     & DAYTON & 0.139 & \textbf{0.093} & \textbf{0.093} & 0.139 & 0.094 & 0.094 & \textbf{0.093} & 0.094 \\
     & DEOK & 0.149 & 0.114 & \textbf{0.113} & 0.149 & 0.117 & 0.117 & 0.117 & 0.117 \\
     & DOM & 0.149 & 0.106 & \textbf{0.105} & 0.150 & 0.108 & 0.108 & 0.112 & 0.115 \\
     & DUQ & 0.123 & 0.096 & \textbf{0.095} & 0.123 & 0.096 & 0.097 & 0.096 & 0.097 \\
     & EKPC & 0.195 & 0.163 & 0.162 & 0.196 & \textbf{0.161} & 0.162 & 0.177 & 0.174 \\
     & FE & 0.126 & \textbf{0.084} & \textbf{0.084} & 0.126 & 0.085 & 0.086 & 0.086 & 0.086 \\
     & NI & 0.112 & \textbf{0.080} & 0.081 & 0.114 & \textbf{0.080} & \textbf{0.080} & 0.083 & 0.081 \\
     & PJME & 0.120 & 0.078 & 0.078 & 0.120 & 0.078 & 0.079 & \textbf{0.077} & 0.079 \\
     & PJMW & 0.150 & \textbf{0.105} & \textbf{0.105} & 0.150 & \textbf{0.105} & \textbf{0.105} & 0.109 & 0.109 \\
     & PJM\_Load & 0.121 & \textbf{0.076} & 0.077 & 0.123 & 0.080 & 0.079 & 0.085 & 0.082 \\
    \cline{1-10}
    \multirow[t]{12}{*}{6} & AEP & 0.270 & \textbf{0.179} & 0.180 & 0.271 & 0.184 & 0.183 & 0.180 & 0.183 \\
     & COMED & 0.264 & 0.174 & 0.177 & 0.265 & 0.179 & 0.180 & 0.173 & \textbf{0.171} \\
     & DAYTON & 0.287 & 0.186 & 0.187 & 0.287 & 0.187 & 0.189 & \textbf{0.184} & 0.188 \\
     & DEOK & 0.310 & \textbf{0.224} & 0.225 & 0.311 & 0.226 & 0.228 & 0.234 & 0.238 \\
     & DOM & 0.305 & \textbf{0.220} & 0.221 & 0.306 & 0.224 & 0.226 & 0.222 & 0.222 \\
     & DUQ & 0.255 & \textbf{0.193} & 0.195 & 0.256 & 0.197 & 0.202 & 0.193 & 0.194 \\
     & EKPC & 0.363 & \textbf{0.288} & 0.290 & 0.364 & 0.291 & 0.291 & 0.305 & 0.311 \\
     & FE & 0.265 & 0.174 & 0.174 & 0.267 & 0.174 & 0.175 & \textbf{0.170} & 0.178 \\
     & NI & 0.249 & 0.160 & 0.163 & 0.250 & 0.164 & 0.165 & \textbf{0.160} & 0.167 \\
     & PJME & 0.261 & 0.167 & 0.167 & 0.262 & 0.173 & 0.174 & \textbf{0.166} & 0.171 \\
     & PJMW & 0.302 & 0.205 & \textbf{0.204} & 0.302 & 0.207 & 0.210 & 0.211 & 0.216 \\
     & PJM\_Load & 0.268 & 0.163 & 0.165 & 0.271 & \textbf{0.163} & 0.165 & 0.167 & 0.177 \\
    \cline{1-10}
    \multirow[t]{12}{*}{12} & AEP & 0.345 & 0.248 & 0.253 & 0.346 & 0.260 & 0.260 & \textbf{0.242} & 0.251 \\
     & COMED & 0.342 & 0.235 & 0.237 & 0.344 & 0.245 & 0.240 & 0.230 & \textbf{0.229} \\
     & DAYTON & 0.372 & 0.250 & 0.256 & 0.373 & 0.257 & 0.261 & \textbf{0.246} & 0.249 \\
     & DEOK & 0.397 & 0.300 & \textbf{0.298} & 0.399 & 0.309 & 0.309 & 0.314 & 0.328 \\
     & DOM & 0.380 & 0.289 & 0.293 & 0.382 & 0.301 & 0.305 & \textbf{0.288} & 0.298 \\
     & DUQ & 0.327 & 0.263 & 0.266 & 0.328 & 0.267 & 0.275 & \textbf{0.252} & 0.252 \\
     & EKPC & 0.434 & 0.376 & 0.381 & 0.436 & \textbf{0.376} & 0.380 & 0.377 & 0.406 \\
     & FE & 0.347 & 0.245 & 0.247 & 0.350 & 0.248 & 0.248 & \textbf{0.232} & 0.236 \\
     & NI & 0.326 & 0.231 & 0.228 & 0.328 & 0.234 & 0.233 & \textbf{0.216} & 0.228 \\
     & PJME & 0.335 & 0.232 & 0.236 & 0.337 & 0.240 & 0.243 & \textbf{0.221} & 0.229 \\
     & PJMW & 0.383 & \textbf{0.275} & 0.279 & 0.384 & 0.280 & 0.285 & 0.277 & 0.278 \\
     & PJM\_Load & 0.355 & 0.231 & 0.235 & 0.360 & 0.243 & 0.241 & \textbf{0.224} & 0.253 \\
    \cline{1-10}
    \bottomrule
    \end{tabular}
}
\caption{Average RMSE scores over 5 runs}
\label{tab:rmse_scores}
\end{table}
\begin{table}[H]
\centering
\resizebox{1\textwidth}{!}{
\begin{tabular}{llrrrrrrrr}
\toprule
 & model & LinearRegression & STAN-3000-3 & STAN-3000-4 & Linear & MLP-3000-3 & MLP-3000-4 & GRU-300-3 & LSTM-300-3 \\
steps ahead & dataset &  &  &  &  &  &  &  &  \\
\midrule
\multirow[t]{12}{*}{1} & AEP & 0.052 & 39.165 & 49.997 & 5.542 & 34.110 & 45.373 & 260.729 & 292.765 \\
 & COMED & 0.033 & 22.634 & 30.677 & 5.041 & 18.381 & 23.557 & 137.853 & 129.579 \\
 & DAYTON & 0.044 & 36.825 & 51.474 & 5.797 & 32.013 & 37.467 & 241.887 & 253.660 \\
 & DEOK & 0.019 & 16.963 & 24.232 & 4.818 & 14.937 & 19.723 & 145.713 & 127.716 \\
 & DOM & 0.045 & 38.394 & 52.931 & 4.835 & 28.730 & 35.949 & 194.783 & 227.935 \\
 & DUQ & 0.047 & 36.757 & 46.997 & 6.077 & 32.257 & 40.539 & 285.547 & 250.858 \\
 & EKPC & 0.015 & 17.803 & 20.755 & 3.627 & 10.953 & 13.969 & 125.131 & 145.141 \\
 & FE & 0.019 & 22.164 & 31.297 & 4.432 & 18.345 & 24.732 & 181.022 & 144.315 \\
 & NI & 0.021 & 20.373 & 22.003 & 4.643 & 19.571 & 25.438 & 136.240 & 150.468 \\
 & PJME & 0.058 & 42.463 & 61.241 & 7.125 & 40.084 & 49.479 & 350.083 & 352.547 \\
 & PJMW & 0.055 & 52.229 & 62.961 & 5.864 & 36.992 & 44.933 & 266.955 & 314.759 \\
 & PJM\_Load & 0.012 & 18.872 & 27.224 & 3.852 & 13.363 & 19.197 & 149.754 & 120.101 \\
\cline{1-10}
\multirow[t]{12}{*}{6} & AEP & 0.055 & 32.104 & 41.375 & 6.817 & 24.253 & 33.230 & 162.993 & 168.799 \\
 & COMED & 0.027 & 19.812 & 27.830 & 6.484 & 14.205 & 19.145 & 80.190 & 99.790 \\
 & DAYTON & 0.050 & 32.234 & 37.205 & 7.247 & 24.602 & 36.347 & 148.397 & 164.175 \\
 & DEOK & 0.023 & 15.657 & 22.686 & 4.642 & 12.377 & 16.656 & 68.340 & 83.693 \\
 & DOM & 0.050 & 29.655 & 41.324 & 6.330 & 20.484 & 25.950 & 122.826 & 123.020 \\
 & DUQ & 0.049 & 29.639 & 40.996 & 7.554 & 21.927 & 31.093 & 122.160 & 148.149 \\
 & EKPC & 0.020 & 11.581 & 15.095 & 5.118 & 7.936 & 10.889 & 46.142 & 49.740 \\
 & FE & 0.029 & 22.277 & 27.349 & 5.118 & 15.845 & 18.175 & 88.635 & 96.380 \\
 & NI & 0.025 & 18.668 & 26.072 & 5.355 & 12.746 & 16.825 & 84.213 & 96.291 \\
 & PJME & 0.064 & 47.394 & 56.034 & 8.565 & 29.769 & 44.394 & 213.840 & 223.237 \\
 & PJMW & 0.063 & 39.004 & 49.894 & 6.981 & 27.673 & 40.770 & 162.749 & 202.707 \\
 & PJM\_Load & 0.014 & 15.156 & 17.521 & 3.819 & 11.133 & 14.188 & 65.671 & 64.076 \\
\cline{1-10}
\multirow[t]{12}{*}{12} & AEP & 0.123 & 27.556 & 36.571 & 6.379 & 21.978 & 27.785 & 133.226 & 174.108 \\
 & COMED & 0.062 & 14.715 & 20.643 & 6.067 & 12.679 & 13.632 & 84.436 & 93.864 \\
 & DAYTON & 0.085 & 30.777 & 35.866 & 8.498 & 22.473 & 30.515 & 169.442 & 190.313 \\
 & DEOK & 0.059 & 14.091 & 17.733 & 4.166 & 10.265 & 12.392 & 60.674 & 73.078 \\
 & DOM & 0.109 & 23.647 & 31.906 & 6.600 & 17.832 & 22.418 & 114.274 & 144.410 \\
 & DUQ & 0.097 & 26.420 & 33.691 & 7.835 & 18.345 & 27.013 & 106.474 & 143.213 \\
 & EKPC & 0.053 & 8.632 & 10.991 & 4.625 & 6.967 & 8.422 & 46.890 & 46.611 \\
 & FE & 0.060 & 19.139 & 25.382 & 5.244 & 12.288 & 16.969 & 122.893 & 148.960 \\
 & NI & 0.056 & 18.618 & 21.029 & 6.250 & 12.673 & 15.215 & 81.129 & 100.758 \\
 & PJME & 0.112 & 39.256 & 45.901 & 8.121 & 25.291 & 32.261 & 191.145 & 207.213 \\
 & PJMW & 0.098 & 30.259 & 41.056 & 7.586 & 23.442 & 32.837 & 146.113 & 171.954 \\
 & PJM\_Load & 0.042 & 11.865 & 15.141 & 3.380 & 7.762 & 9.215 & 59.248 & 73.959 \\
\cline{1-10}
\bottomrule
\end{tabular}
}
\caption{Average Training Time (seconds)}
\label{tab:time_scores}
\end{table}
\begin{table}[H]
\centering
\resizebox{1\textwidth}{!}{
\begin{tabular}{l|cccccccc}
\toprule
steps\_ahead & LinearRegression & STAN-3000-3 & STAN-3000-4 & Linear & MLP-3000-3 & MLP-3000-4 & GRU-300-3 & LSTM-300-3 \\
\midrule
1 & 46 & 18.2M & 27.2M & 46 & 18.1M & 27.2M & 1.37M & 1.82M \\
6 & 51 & 18.2M & 27.2M & 276 & 18.2M & 27.2M & 1.44M & 1.89M \\
12 & 72 & 18.3M & 27.3M & 732 & 18.2M & 27.2M & 1.57M & 2.02M \\
\bottomrule
\end{tabular}
}
\caption{Number of trainable parameters for each model architecture and prediction horizon}
\label{tab:parameter_counts}
\end{table}

\section{Conclusion}
In this paper, we introduced Smooth Transition Autoregressive Networks (STAN), a neural architecture designed to bridge the gap between classical econometric models and modern deep learning approaches. Our work demonstrates how the theoretical foundations of traditional econometric models, particularly STAR models, can be successfully integrated into the deep learning framework. This integration represents a step forward in bringing together the rich history of econometric modeling with the flexibility and power of neural networks. Our empirical evaluation on the PJM Hourly Energy Consumption dataset reveals STAN's distinctive performance profile: superior accuracy in short-term forecasting across most regions, with competitive performance at longer horizons where recurrent architectures take the lead. STAN achieves this while maintaining training efficiency comparable to MLPs and significantly faster than recurrent networks, despite its parameter count being similar to MLPs and larger than recurrent architectures. This efficiency stems from STAN's feed-forward structure, which preserves the computational advantages of MLPs while incorporating regime-switching capabilities. While effective, our current implementation presents opportunities for enhancement. The univariate design could be extended to multivariate scenarios, potentially capturing richer dependencies in the data. Furthermore, formalizing the theoretical connections between STAN's transition mechanism and traditional STAR models could provide deeper insights into its behavior and guide future improvements. Future research directions include extending STAN to multivariate settings, developing interpretability methods for learned regime transitions, and exploring applications in other domains where regime-switching behavior is prevalent. The success of STAN demonstrates the potential of synthesizing econometric theory with deep learning, suggesting a pathway toward models that combine classical interpretability with modern architectural advances. These results encourage further development of hybrid approaches that leverage both economic theory and machine learning innovations. Such integration not only advances modeling capabilities but also helps unify traditional econometrics and contemporary deep learning, working toward more robust and interpretable solutions for economic and financial applications.

\bibliography{bib}

\begin{appendices}
\begin{table}[H]
\centering
\resizebox{\textwidth}{!}{
\begin{tabular}{llllllllll}
\toprule
 & model & LinearRegression & STAN-3000-3 & STAN-3000-4 & Linear & MLP-3000-3 & MLP-3000-4 & GRU-300-3 & LSTM-300-3 \\
steps ahead & dataset &  &  &  &  &  &  &  &  \\
\midrule
\multirow[t]{12}{*}{1} & AEP & \textbf{0.000} & 3.801 & 6.238 & 0.307 & 3.021 & 2.990 & 5.312 & 3.945 \\
 & COMED & \textbf{0.000} & 6.638 & 6.879 & 3.556 & 4.603 & 5.671 & 3.383 & 7.106 \\
 & DAYTON & \textbf{0.000} & 3.778 & 4.593 & 0.169 & 4.840 & 4.607 & 3.492 & 9.202 \\
 & DEOK & \textbf{0.000} & 10.079 & 3.778 & 0.904 & 9.697 & 3.865 & 6.282 & 22.146 \\
 & DOM & \textbf{0.000} & 3.855 & 4.794 & 0.371 & 7.647 & 9.701 & 14.471 & 18.321 \\
 & DUQ & \textbf{0.000} & 3.730 & 4.492 & 0.509 & 3.654 & 3.662 & 4.994 & 7.660 \\
 & EKPC & \textbf{0.000} & 33.374 & 22.251 & 1.629 & 9.897 & 13.739 & 16.990 & 61.610 \\
 & FE & \textbf{0.000} & 4.675 & 5.323 & 0.856 & 7.916 & 10.278 & 5.634 & 10.551 \\
 & NI & \textbf{0.000} & 5.472 & 8.293 & 5.247 & 4.750 & 4.562 & 17.663 & 25.753 \\
 & PJME & \textbf{0.000} & 1.892 & 5.127 & 0.426 & 3.084 & 3.053 & 5.815 & 4.628 \\
 & PJMW & \textbf{0.000} & 6.070 & 3.460 & 0.525 & 4.839 & 5.284 & 4.153 & 9.427 \\
 & PJM\_Load & \textbf{0.000} & 6.461 & 4.960 & 2.909 & 12.298 & 5.867 & 10.737 & 7.751 \\
\cline{1-10}
\multirow[t]{12}{*}{6} & AEP & \textbf{0.000} & 6.084 & 7.422 & 1.224 & 9.906 & 12.638 & 16.876 & 13.548 \\
 & COMED & \textbf{0.000} & 13.453 & 15.871 & 3.107 & 12.696 & 23.688 & 8.168 & 19.918 \\
 & DAYTON & \textbf{0.000} & 3.293 & 16.763 & 0.378 & 7.477 & 23.791 & 17.840 & 37.236 \\
 & DEOK & \textbf{0.000} & 14.259 & 18.002 & 5.250 & 16.522 & 8.402 & 32.288 & 29.376 \\
 & DOM & \textbf{0.000} & 18.203 & 15.411 & 1.991 & 39.761 & 25.215 & 19.336 & 13.424 \\
 & DUQ & \textbf{0.000} & 9.043 & 18.384 & 1.839 & 14.694 & 4.654 & 8.753 & 22.338 \\
 & EKPC & \textbf{0.000} & 7.844 & 10.037 & 4.554 & 37.759 & 13.828 & 19.960 & 53.200 \\
 & FE & \textbf{0.000} & 14.248 & 7.279 & 4.076 & 11.854 & 18.949 & 24.939 & 13.852 \\
 & NI & \textbf{0.000} & 9.170 & 7.215 & 2.826 & 11.620 & 22.351 & 14.603 & 8.296 \\
 & PJME & \textbf{0.000} & 11.432 & 15.592 & 0.846 & 8.375 & 15.951 & 33.149 & 17.014 \\
 & PJMW & \textbf{0.000} & 12.820 & 6.019 & 1.069 & 5.814 & 21.433 & 17.313 & 25.700 \\
 & PJM\_Load & \textbf{0.000} & 14.505 & 8.330 & 7.297 & 6.544 & 16.824 & 26.262 & 23.334 \\
\cline{1-10}
\multirow[t]{12}{*}{12} & AEP & \textbf{0.000} & 21.291 & 16.370 & 1.898 & 9.831 & 39.677 & 18.638 & 37.362 \\
 & COMED & \textbf{0.000} & 14.792 & 23.897 & 6.121 & 35.851 & 30.968 & 66.296 & 31.127 \\
 & DAYTON & \textbf{0.000} & 1.239 & 17.874 & 1.439 & 39.666 & 41.496 & 35.829 & 24.556 \\
 & DEOK & \textbf{0.000} & 39.592 & 8.925 & 2.407 & 42.330 & 21.599 & 31.041 & 40.757 \\
 & DOM & \textbf{0.000} & 14.776 & 22.199 & 2.401 & 55.409 & 53.144 & 18.345 & 20.340 \\
 & DUQ & \textbf{0.000} & 15.854 & 36.618 & 2.961 & 38.637 & 32.282 & 27.375 & 31.708 \\
 & EKPC & \textbf{0.000} & 36.847 & 28.656 & 4.648 & 37.328 & 58.628 & 32.404 & 164.720 \\
 & FE & \textbf{0.000} & 13.123 & 21.025 & 4.810 & 28.726 & 14.839 & 34.276 & 36.102 \\
 & NI & \textbf{0.000} & 16.504 & 12.739 & 4.041 & 16.994 & 27.027 & 28.155 & 34.456 \\
 & PJME & \textbf{0.000} & 4.454 & 7.018 & 4.791 & 14.908 & 9.222 & 11.758 & 30.881 \\
 & PJMW & \textbf{0.000} & 23.352 & 21.308 & 0.640 & 13.596 & 43.129 & 28.983 & 36.114 \\
 & PJM\_Load & \textbf{0.000} & 25.217 & 17.718 & 7.267 & 43.090 & 26.442 & 49.481 & 32.014 \\
\cline{1-10}
\bottomrule
\end{tabular}
}
\caption{Standard Deviation of the RMSE over runs (multiplied by 100)}
\label{tab:rmse_std}
\end{table}
\begin{table}[H]
\centering
\resizebox{\textwidth}{!}{
\begin{tabular}{llllllllll}
\toprule
 & model & LinearRegression & STAN-3000-3 & STAN-3000-4 & Linear & MLP-3000-3 & MLP-3000-4 & GRU-300-3 & LSTM-300-3 \\
steps ahead & dataset &  &  &  &  &  &  &  &  \\
\midrule
\multirow[t]{12}{*}{1} & AEP & 0.004 & 7.201 & 7.333 & 0.834 & 1.207 & 4.781 & 33.774 & 29.262 \\
 & COMED & 0.023 & 2.879 & 3.257 & 0.328 & 3.182 & 2.354 & 12.110 & 9.334 \\
 & DAYTON & 0.003 & 4.458 & 4.342 & 0.447 & 3.181 & 3.418 & 25.941 & 31.649 \\
 & DEOK & 0.002 & 1.433 & 2.168 & 0.276 & 1.557 & 1.331 & 17.548 & 20.308 \\
 & DOM & 0.002 & 4.210 & 7.591 & 0.665 & 2.729 & 3.561 & 21.129 & 28.434 \\
 & DUQ & 0.005 & 3.232 & 4.785 & 0.856 & 4.318 & 3.836 & 40.466 & 18.504 \\
 & EKPC & 0.001 & 2.408 & 1.259 & 0.213 & 1.061 & 1.584 & 18.987 & 14.842 \\
 & FE & 0.001 & 2.911 & 3.355 & 0.530 & 1.478 & 4.471 & 15.914 & 21.002 \\
 & NI & 0.002 & 2.691 & 2.644 & 0.555 & 2.412 & 3.008 & 27.783 & 50.897 \\
 & PJME & 0.003 & 3.972 & 4.380 & 0.576 & 2.650 & 3.468 & 18.398 & 66.632 \\
 & PJMW & 0.002 & 4.600 & 8.296 & 0.535 & 4.058 & 3.884 & 14.621 & 20.803 \\
 & PJM\_Load & 0.000 & 3.553 & 2.859 & 0.541 & 1.852 & 2.612 & 27.310 & 16.188 \\
\cline{1-10}
\multirow[t]{12}{*}{6} & AEP & 0.001 & 3.672 & 3.763 & 0.639 & 2.333 & 2.777 & 20.095 & 11.504 \\
 & COMED & 0.004 & 0.949 & 2.359 & 0.616 & 1.372 & 3.560 & 4.384 & 7.123 \\
 & DAYTON & 0.002 & 3.531 & 5.343 & 0.601 & 2.132 & 2.251 & 20.193 & 16.286 \\
 & DEOK & 0.002 & 1.421 & 1.372 & 0.461 & 1.953 & 0.650 & 4.654 & 5.076 \\
 & DOM & 0.002 & 1.132 & 4.255 & 0.716 & 1.943 & 2.948 & 8.263 & 16.126 \\
 & DUQ & 0.001 & 1.392 & 2.396 & 0.647 & 1.479 & 1.930 & 8.763 & 12.172 \\
 & EKPC & 0.003 & 1.008 & 1.444 & 0.991 & 0.675 & 0.894 & 3.436 & 5.280 \\
 & FE & 0.005 & 2.496 & 3.991 & 0.522 & 2.361 & 2.665 & 4.581 & 9.090 \\
 & NI & 0.002 & 1.696 & 4.620 & 0.399 & 0.695 & 1.199 & 9.096 & 2.757 \\
 & PJME & 0.001 & 5.305 & 3.404 & 1.122 & 1.875 & 4.798 & 26.206 & 20.900 \\
 & PJMW & 0.002 & 3.251 & 3.294 & 1.272 & 1.602 & 3.033 & 12.282 & 13.728 \\
 & PJM\_Load & 0.001 & 2.445 & 1.183 & 0.231 & 0.758 & 1.098 & 8.278 & 1.599 \\
\cline{1-10}
\multirow[t]{12}{*}{12} & AEP & 0.033 & 2.419 & 3.173 & 0.487 & 1.796 & 3.539 & 8.538 & 12.732 \\
 & COMED & 0.003 & 1.728 & 1.663 & 1.080 & 1.611 & 0.999 & 13.769 & 6.189 \\
 & DAYTON & 0.006 & 1.385 & 5.333 & 1.035 & 3.721 & 4.834 & 16.709 & 29.429 \\
 & DEOK & 0.008 & 0.692 & 1.330 & 0.298 & 1.280 & 1.508 & 9.533 & 6.269 \\
 & DOM & 0.024 & 1.002 & 1.205 & 1.303 & 2.184 & 1.370 & 11.340 & 13.800 \\
 & DUQ & 0.018 & 3.014 & 2.011 & 1.116 & 0.724 & 1.930 & 7.255 & 11.026 \\
 & EKPC & 0.011 & 0.695 & 0.443 & 0.429 & 1.233 & 0.590 & 3.450 & 3.860 \\
 & FE & 0.007 & 1.587 & 2.699 & 0.585 & 0.953 & 1.177 & 22.719 & 25.915 \\
 & NI & 0.010 & 1.761 & 1.044 & 0.485 & 1.182 & 1.828 & 7.226 & 9.105 \\
 & PJME & 0.021 & 2.547 & 1.527 & 1.473 & 2.045 & 0.634 & 16.232 & 23.398 \\
 & PJMW & 0.006 & 1.974 & 2.583 & 0.509 & 1.556 & 2.889 & 8.161 & 10.636 \\
 & PJM\_Load & 0.009 & 1.569 & 1.494 & 0.385 & 0.592 & 0.446 & 6.067 & 5.103 \\
\cline{1-10}
\bottomrule
\end{tabular}
}
\caption{Standard Deviation of the Training time}
\label{tab:time_std}
\end{table}

\end{appendices}
\end{document}